\pdfoutput=1
\documentclass[11pt]{article}
\usepackage[final]{acl}
\usepackage{times}
\usepackage{latexsym}
\usepackage[T1]{fontenc}
\usepackage[utf8]{inputenc}
\usepackage{microtype}

\usepackage{graphicx} 

\usepackage{algorithm}
\usepackage{algorithmic}

\usepackage{threeparttable}
\usepackage{booktabs}
\usepackage{multirow}
\usepackage{multicol}
\usepackage{makecell}
\usepackage{amsmath}
\usepackage{newfloat}
\usepackage{listings}
\usepackage{rotating}

\usepackage{enumitem}

\title{Template-based Approach to Zero-shot Intent Recognition}
\author{
    ~\textbf{Dmitry Lamanov}\textsuperscript{1},
    ~\textbf{Pavel Burnyshev}\textsuperscript{1},
    ~\textbf{Ekaterina Artemova}\textsuperscript{1,2}, \\
    ~\textbf{Valentin Malykh}\textsuperscript{1},
    ~\textbf{Andrey Bout}\textsuperscript{1},
    ~\textbf{Irina Piontkovskaya}\textsuperscript{1} \\
	\textsuperscript{1}Huawei Noah’s Ark lab,
	\textsuperscript{2}HSE University \\ \\
	
	  \small{
    \textbf{Correspondence:} \href{mailto: piontkovskaya.irina@huawei.com}{piontkovskaya.irina@huawei.com}}
}
\begin{document}

\maketitle

\begin{abstract}
The recent advances in transfer learning techniques and pre-training of large contextualized encoders foster innovation in real-life applications, including dialog assistants. Practical needs of intent recognition require effective data usage and the ability to constantly update supported intents, adopting new ones, and abandoning outdated ones. In particular, the generalized zero-shot paradigm, in which the model is trained on the seen intents and tested on both seen and unseen intents, is taking on new importance.
In this paper, we explore the generalized zero-shot setup for intent recognition. Following best practices for zero-shot text classification, we treat the task with a sentence pair modeling approach. We outperform previous state-of-the-art f1-measure by up to 16\% for unseen intents, using intent labels and user utterances and without accessing external sources (such as knowledge bases). Further enhancement includes lexicalization of intent labels, which improves performance by up to 7\%.  By using task transferring from other sentence pair tasks, such as Natural Language Inference, we gain additional improvements. 
\end{abstract}

\section{Introduction}

User intent recognition is one of the key components of dialog assistants. With the advent of deep learning models, deep classifiers have been used throughout to recognize user intents. A common setup for the task \cite{chen2019bert,wu2020tod,casanueva2020efficient} involves an omnipresent pre-trained language model \cite{devlin2018bert,liu2019roberta,sanh2019distilbert}, equipped with a classification head, learned to predict intents. However, if the dialog assistant is extended with new skills or applications, new intents may appear. In this case, the intent recognition model needs to be re-trained. In turn, re-training the model requires annotated data, the scope of which is inherently limited. Hence, handling unseen events defies the common setup and poses new challenges. To this end, {\bf generalized zero-shot (GZS)} learning scenario \cite{xian2018zero}, in which the model is presented at the training phase with  {\it seen} intents and at the inference phase with both {\it seen} and {\it unseen} intents, becomes more compelling and relevant for real-life setups.  The main challenge lies in developing a model capable of processing  {\it seen} and {\it unseen} intents at comparable performance levels.

Recent frameworks for GZS intent recognition are designed as complex multi-stage pipelines, which involve: detecting unseen intents \cite{yan2020unknown}, learning intent prototypes \cite{si2020learning}, leveraging common sense knowledge graphs \cite{siddique2021generalized}. Such architecture choices may appear untrustworthy: using learnable unseen detectors leads to cascading failures; relying on external knowledge makes the framework hardly adjustable to low-resource domains and languages. Finally, interactions between different framework's components may be not transparent, so it becomes difficult to trace back the prediction and guarantee the interpretability of results. 
 
At the same time, recent works in the general domain GZL classification are centered on the newly established approach of  \citet{yin2019benchmarking}, who formulate the task as a textual entailment problem. The class's description is treated as a hypothesis and the text – as a premise. The GZL classification becomes a binary problem: to predict whether the hypothesis entails the premise or not. Entailment-based approaches have been successfully used for information extraction  \cite{haneczok2021fine,lyuzero,sainz2021ask2transformers} and for dataless classification \cite{ma-etal-2021-issues}. However, the entailment-based setup has not been properly explored for GZS intent recognition to the best of our knowledge. 

This paper aims to fill in the gap and extensively evaluate entailment-based approaches for GZS intent recognition. Given a meaningful intent label, such as \texttt{reset\_settings}, and an input utterance, such as \textit{I want my original settings back}, the classifier is trained to predict if the utterance should be assigned with the presented intent or not. To this end, we make use of pre-trained language models, which encode a two-fold input (intent label and an utterance) simultaneously and fuse it at intermediate layers with the help of the attention mechanism. 

We adopt three dialog datasets for GZS intent recognition and show that sentence pair modeling outperforms competing approaches and establishes new state-of-the-art results. Next, we implement multiple techniques, yielding an even higher increase in performance. Noticing that in all datasets considered, most intent labels are either noun or verb phrases, we implement a small set of lexicalizing templates that turn intent labels into plausible sentences. For example, an intent label  \texttt{reset\_settings} is re-written as  \textit{The user wants to reset settings}. Such lexicalized intent labels appear less surprising to the language model than intact intent labels. Hence, lexicalization of intent labels helps the language model to learn correlations between inputs efficiently. Other improvements are based on standard engineering techniques, such as hard example mining and task transferring.

Last but not least, we explore two setups in which even less data is provided by restricting access to various parts of annotated data. First, if absolutely no data is available, we explore strategies for transferring from models pre-trained with natural language inference data. Second, in the dataless setup only seen intent labels are granted and there are no annotated utterances, we seek to generate synthetic data from them by using off-the-shelf models for paraphrasing. We show that the sentence pair modeling approach to GZS intent recognition delivers adequate results, even when trained with synthetic utterances, but fails to transfer from other datasets.

The key contributions of the paper are as follows: 
\begin{enumerate}[noitemsep,nolistsep]
    \item we discover that sentence pair modeling approach to GZS intent recognition establishes new state-of-the-art results;
    \item we show that lexicalization of intent labels yields further significant improvements;
    \item  we use task transferring, training in dataless regime and conduct error analysis to investigate the strengths and weaknesses of sentence pair modeling approach.
    
\end{enumerate}

\section{Related Work}
Our work is related to two lines of research: zero-shot learning with natural language descriptions and intent recognition. We focus on adopting existing ideas for zero-shot text classification to intent recognition.

{\bf Zero-shot learning} has shown tremendous progress in NLP in recent years. The scope of the tasks, studied in GZS setup, ranges from text classification \cite{yin2019benchmarking} to event extraction \cite{haneczok2021fine,lyuzero}, named entity recognition \cite{li2020unified} and entity linking \cite{logeswaran2019zero}. A number of datasets for benchmarking zero-shot methods has been developed. To name a few, \citet{yin2019benchmarking} create a benchmark for general domain text classification.  SGD \cite{DBLP:conf/aaai/RastogiZSGK20} allows for zero-shot intent recognition. 

Recent research has adopted a scope of novel approaches, utilizing natural language descriptions, aimed at zero-shot setup. Text classification can be treated in form of a textual entailment problem \cite{yin2019benchmarking}, in which the model learns to match features from class' description and text, relying on early fusion between inputs inside the attention mechanism. The model can be fine-tuned solely of the task's data or utilize pre-training with textual entailment and natural language inference \cite{sainz2021ask2transformers}.  However,  dataless classification with the help of models, pre-trained for textual entailment only appears problematic due to models' high variance and  instability \cite{ma-etal-2021-issues}. This justifies the rising need for learnable domain transferring \cite{yin2020universal} and  self-training \cite{ye2020zero}, aimed at leveraging unlabeled data and alleviating domain shift between seen and unseen classes. 

{\bf Intent recognition} Supervised intent recognition requires training a classifier with a softmax layer on top. Off-the-shelf pre-trained language models or sentence encoders are used to embed an input utterance, fed further to the classifier \cite{casanueva2020efficient}.  Augmentation techniques help to increase the amount of training data and increase performance \cite{xia2020composed}. Practical needs require the classifier to support emerging intents. Re-training a traditional classifier may turn out resource-greedy and costly. This motivates work in (generalized) zero-shot intent recognition,  i.e. handling  seen and unseen intents simultaneously. Early approaches to GZS intent recognition adopted {\bf capsule networks} to learn low-dimensional representations of intents. IntentCapsNet \cite{xia2018zero} is built upon three capsule modules, organized hierarchically: the lower module extracts semantic features from input utterances. Two upper modules execute recognition of seen and unseen intents independently from each other.  ReCapsNet \cite{liu-etal-2019-reconstructing} is built upon a transformation schema, which detects unseen events and makes predictions based on unseen intents' similarity to the seen ones. SEG \cite{yan2020unknown} utilizes {\bf Gaussian mixture models} to learn intent representations by maximising margins between them.   One of the concurrent approaches, CTIR \cite{si2020learning} (Class-Transductive Intent Representations) learns {\bf intent representations from  intent labels} to model inter-intent connections.  CTIR is not a stand-alone solution but rather integrates existing models, such as BERT, CNN, or CapsNet. The framework expands the prediction space at the training stage to be able to include unseen classes, with the unseen label names serving as pseudo-utterances. The current state-of-the-art performance belongs to RIDE \cite{siddique2021generalized},  an intent detection model that leverages {\bf common knowledge} from ConceptNet. RIDE captures  semantic relationships between utterances and intent labels considering concepts in an utterance linked to those in an intent label. 




\section{Sentence pair modelling for intent recognition}

\subsection{Problem formulation}

Let $\mathcal{X}$ be the set of utterances, $\mathcal{S} = \{y_1, \ldots, y_k\}$ be the set of seen intents and  $\mathcal{U} = \{y_{k+1}, \ldots, y_n\}$ be the set of unseen intents. The training data consists of annotated utterances $\{x_i, y_j\}$. At the test time, the model is presented with a new utterance.  In the GZS setup the model chooses an intent from both seen and unseen $y_j \in \mathcal{S} \cup \mathcal{U}$.


\subsection{Our approach}

A contextualized encoder is trained to make a binary prediction: whether the utterance~$x_i$ is assigned with the intent~$y_j$ or not. The model encodes the intent description and the utterance, concatenated by the separation token \texttt{[SEP]}. The representation of the \texttt{[CLS]} token is fed into a classification head, which makes the desired prediction $P(1|y_j, x_i)$. This approach follows standard sentence pair (SP) modeling setup.

Given an intent $y_j$, the model is trained to make a positive prediction for an in-class utterance $x_i^+$ and a negative prediction for an out-of-class utterance $x_i^-$, sampled from another intent. At the train time, the model is trained with seen intents only $y_j \in \mathcal{S}$.

On the test time, given an utterance $x_i^{test}$, we loop over all intents $y_j \in \mathcal{S} \cup \mathcal{U}$ and record the probability of the positive class. Finally, we assign to the utterance $x_i^{test}$ such $y^{*}$, that provides with the maximum probability of the positive class:

\[y^{*} = \operatorname*{arg\,max}_{y_j \in \mathcal{S} \cup \mathcal{U}} P(1|y_j, x_i^{test}) \]


\paragraph{Contextualized encoders.} We use RoBERTa$_{base}$~\cite{liu2019roberta} as the main and default contextualized encoder in our experiments, as it shows superior performance to BERT~\cite{devlin2018bert} in many downstream applications. RoBERTa's distilled version, DistilRoBERTa~\cite{sanh2019distilbert} is used to evaluate lighter, less computationally expensive models. Also, we use a pre-trained task-oriented dialogue model, TOD-BERT~\cite{wu2020tod} to evaluate whether domain models should be preferred. 

We used models, released by HuggingFace library \cite{wolf2020transformers}: \texttt{roberta-base}, \texttt{distilroberta-base} and  \texttt{TODBERT/TOD-BERT-JNT-V1} .

\setlength{\textfloatsep}{0.1cm}
\begin{table}[!t]
\centering
\begin{tabular}{lp{5cm}}
\toprule
ID &  Template  \\
\midrule
& declarative templates \\
\midrule
d$_1$ & {\it the user wants to } \\
& the user wants to book a hotel \\ 
d$_2$ & {\it tell the user how to} \\
& tell the user how to book a hotel \\
\midrule
& question templates \\
\midrule
q$_1$ & {\it does the user want to} \\
& does the user want to book a hotel \\
q$_2$ & {\it how do I} \\
& how do I book a hotel \\
\bottomrule
\end{tabular}
\caption{Lexicalization templates, applied to intent labels. Examples are provided for the intent label ``book hotel''. }
\label{tab:patterns}
\end{table}


\paragraph{Negative sampling strategies} include (i) sampling negative utterances for a fixed intent, denoted as $(y_j$, $x_i^+), (y_j$, $x_i^-)$; (ii) sampling negative intents for a fixed utterance, denoted as $(y_j^{+}$, $x_i), (y_j^{-}$, $x_i)$.

Both strategies support sampling with hard examples. In the first case (i), we treat an utterance $x_i^-$ as a hard negative one for intent $y_j$, if there exists such in-class utterance $x_i^+$, so that the similarity between $x_i^+$ and $x_i^-$ is higher than a predefined threshold. To this end, to compute semantic similarity, we make use of SentenceBERT \cite{reimers2019sentence} cosine similarity. For a given positive in-class utterance, we selected the top-100 most similar negative out-of-class utterance based on the values of cosine similarity. In the second case (ii), we use the same approach to sample hard negative intents $y_j^-$, given an utterance  $x_i$, assigned with the positive intent $y_j^+$. Again, we compute semantic similarity between intent labels and sample an intent $y_j^-$ with probability based on similarity score with intent $y_j^+$.  To justify the need to sample hard negative examples, we experiment with random sampling, choosing randomly (iii) negative utterances or (iv) negative intents.


\paragraph{Lexicalization of intent labels} utilizes simple grammar templates to convert intent labels into natural-sounding sentences. For this aim, we utilize two types of templates: (i)~declarative templates ({\it ``the user wants to''}) and  (ii)~question templates ({\it ``does the user want to''}). Most intent labels  take a form of a verb phrase (VERB + NOUN$^{+}$), such as \texttt{book\_hotel}  or a noun phrase (NOUN$^{+}$), such as \texttt{flight\_status}. We develop the set of rules that parses an intent label, detects whether it is a verb phrase or a noun phrase\footnote{We use a basic NLTK POS tagger to process intent labels.}, and lexicalizes it using one of the templates using the following expression: {\it template} + VERB + {\it a/an} + NOUN$^{+}$. If the intent label is recognized as a noun phrase, the VERB slot is filled with an auxiliary verb, ``get''. This way, we achieve such sentences: \textit{the user wants to book a hotel} and \textit{does the user want to get a flight status}. The templates implemented are shown in Table~\ref{tab:patterns}.

Lexicalization templates were constructed from the most frequent utterance prefixes, computed for all datasets. This way, lexicalized intents sound natural and are close to the real utterances.  We use declarative and question templates because the datasets consist of such utterance types.  We experimented with a larger number of lexicalization templates, but as there is no significant difference in performance, we limited ourselves to two templates of each kind for the sake of brevity. 
\begin{table*}[!ht] 
\centering
\begin{threeparttable}
\resizebox{\textwidth}{!}{%
\begin{tabular}{ccccccccccccc}
\toprule
\multirow{3}{*}{Method} & 
\multicolumn{4}{c}{SGD} &
\multicolumn{4}{c}{MultiWoZ} &
\multicolumn{4}{c}{CLINC} \\
\cmidrule{2-13}
& \multicolumn{2}{c}{Unseen}&
\multicolumn{2}{c}{Seen}&
\multicolumn{2}{c}{Unseen}&
\multicolumn{2}{c}{Seen}&
\multicolumn{2}{c}{Unseen}&
\multicolumn{2}{c}{Seen} \\
\cmidrule{2-13}
 & Acc & F1 & Acc & F1 & Acc & F1 
 & Acc & F1 & Acc & F1 & Acc & F1 \\
\midrule
SEG& 0.372 & 0.403 & 0.613 & 0.636 & 0.371 & 0.414 & 0.652 & 0.646&
-& -& -& - \\
RIDE+PU& 0.590 & 0.573 & 0.832 & 0.830 & 0.569 & 0.521 & 0.884 & 0.885 &
\bf{0.798} & 0.573 & 0.908 & 0.912 \\

ZSDNN + CTIR&
    0.603 &     0.580 &   0.809 &      0.878 &
    0.468 &        0.437 &   0.827 &      0.892  &
    0.561 &        0.493 &   0.904 &      0.871 \\
CapsNet + CTIR & 
    0.567  &        0.507  &   0.897  &      0.912 &
    0.481  &        0.404  &   0.903 &      0.906  &
    0.530  &        0.572  &   0.866  &      0.883 \\
\midrule
SP RoBERTa (ours)& 
0.698 & 0.732 & 0.917 & 0.925
& 0.606 &        0.686 &   0.903 &      0.919 
& 0.661 &        0.742  &   \bf{0.946} &      \bf{0.954}\\
\makecell{SP RoBERTa\\ + templates (ours)}& 
\bf{0.750} &     \bf{0.805} &   \bf{0.931} &   \bf{0.934} &
\bf{0.624}&	\bf{0.722}&	\bf{0.941}&	\bf{0.948}&
0.692 &     \bf{0.766} &   0.927 &   0.931 \\
\bottomrule
\end{tabular}
}
\caption{Comparison of different methods. SP stands for Sentence Pair modeling approach. SP RoBERTa (ours) shows consistent improvements of F1 across all datasets for seen and unseen intents. The usage of lexicalized templates improves performance. }

\label{tab:all_results}
\end{threeparttable}
\end{table*}

\paragraph{Task transferring} Task transferring from other tasks to GZS intent recognition allows to  estimate whether (i)  pre-trained task-specific models can be used without any additional fine-tuning, reducing the need of annotated data and (ii)  pre-training on other tasks and further fine-tuning is beneficial for the final performance. 

There are multiple tasks and fine-tuned contextualized encoders, which we may exploit for task transferring experiments. For the sake of time and resources,  we did not fine-tune any models on our own, but rather adopted a few suitable models from HuggigngFace library, which were fine-tuned on the Multi-Genre Natural Language Inference (MultiNLI)  dataset \cite{N18-1101}:   BERT-NLI (\texttt{textattack/bert-base-uncased-MNLI}), BART-NLI (\texttt{bart-large-mnli}), RoBERTa-NLI (\texttt{textattack/roberta-base-MNLI}).

\paragraph{Dataless classification} We experiment with a dataless classification scenario, in which we train the models on synthetic data. To this end, we used three pre-trained three paraphrasing models to paraphrase lexicalized intent labels. For example, the intent label \texttt{get alarms} is first lexicalized as \textit{tell the user how to get alarms} and then paraphrased as \textit{What’s the best way to get an alarm?}. Next, we merge all sentences, paraphrased with different models, into a single training set. Finally, we train the GZS model with the lexicalized intent labels and their paraphrased versions without using any annotated utterances. 

The T5-based \cite{2020t5} and Pegasus-based \cite{zhang2020pegasus} paraphrasers (\texttt{ramsrigouthamg/t5\_paraphraser}, \texttt{Vamsi/T5\_Paraphrase\_Paws} \texttt{tuner007/pegasus\_paraphrase}) were adopted from the HuggigngFace library and were used with default parameters and beam size equal to 25. 

\section{Datasets}

\paragraph{SGD} (Schema-Guided Dialog) \cite{DBLP:conf/aaai/RastogiZSGK20}  contains dialogues from 16 domains and 46 intents and provides the explicit train/dev/test split, aimed at the GZSL setup. Three domains are available only in the test set. This is the only dataset, providing short intent descriptions, which we use instead of intent labels.  To pre-process the SGD dataset, we keep utterances where users express an intent, selecting utterances in one of the two cases: (i) first utterances in the dialogue and (ii) an utterance that changes the dialogue state and expresses a new intent. 
We use pre-processed utterances from original train/dev/test sets for the GZS  setup directly without any additional splitting.

\paragraph{MultiWoZ 2.2} (Multi-domain Wizard of Oz) \cite{budzianowski-etal-2018-multiwoz}  is treated same way as SGD: we keep utterances that express an intent and we get 27.5K utterances, spanning over 11 intents from 7 different domains. We used 8 (out of 11) randomly selected intents as seen for training. 30\% utterances from seen intents. All utterances implying unseen intents are used for testing. Test utterances for seen intents are sampled in a stratified way, based on their support in the original dataset.


\paragraph{CLINC} \cite{larson-etal-2019-evaluation}  contains 23,700 utterances, of which 22,500 cover 150 in-scope intents, grouped into ten domains. We follow the standard practice to randomly select 3/4 of the in-scope intents as seen (112 out of 150) and 1/4 as unseen (38 out of 150). The random split was made the same way as for MultiWoZ.

\section{Experiments}
\begin{table*}[ht]
\centering
\begin{threeparttable}
\resizebox{\textwidth}{!}{%
\begin{tabular}{ccccccc}
\toprule
\multirow{3}{*}{Method} & 
\multicolumn{2}{c}{SGD} &
\multicolumn{2}{c}{MultiWoZ} &
\multicolumn{2}{c}{CLINC} \\
 \cmidrule{2-7}
 & Acc & F1 & Acc & F1 & Acc & F1 \\
\midrule
SP RoBERTa &\bf{0.687} $\pm$ 0.018 & 0.716 $\pm$ 0.016
& 0.594  $\pm$  0.180 & 0.705   $\pm$ 0.157
& \bf{0.639}  $\pm$  0.038 & \bf{0.731}  $\pm$  0.028 \\
SP BERT & 0.668 $\pm$ 0.001 & 0.701 $\pm$ 0.001 
& 0.604 $\pm$ 0.190 & 0.704 $\pm$ 0.162 
& 0.613 $\pm$ 0.023 & 0.694 $\pm$ 0.031 \\
SP TOD-BERT & 0.658 $\pm$ 0.055 & \bf{0.724} $\pm$ 0.042
& \bf{0.629} $\pm$ 0.235 & \bf{0.715} $\pm$ 0.241
& 0.625 $\pm$ 0.029 & 0.704 $\pm$ 0.034 \\
SP DistilRoBERTa & 0.658 $\pm$ 0.046 & 0.710 $\pm$ 0.022 & 0.603 $\pm$ 0.208 & 0.701 $\pm$ 0.213 & 0.583 $\pm$ 0.030 & 0.672 $\pm$ 0.029 \\
\midrule
SP RoBERTa + random IS
& 0.687 $\pm$ 0.018 & 0.716 $\pm$ 0.016
& 0.594 $\pm$ 0.180 & \bf{0.705} $\pm$ 0.157 
& 0.639 $\pm$ 0.038 & 0.731 $\pm$ 0.028 \\
SP RoBERTa +  random US
& 0.677 $\pm$ 0.017 & 0.707 $\pm$ 0.014 
& 0.531 $\pm$ 0.218 & 0.632 $\pm$ 0.217
& 0.658 $\pm$ 0.043 & 0.735 $\pm$ 0.036 \\
SP RoBERTa +  hard IS
& \bf{0.741} $\pm$ 0.010 & \bf{0.786} $\pm$ 0.017 
& 0.561 $\pm$ 0.177 & 0.680 $\pm$ 0.136 
& 0.590 $\pm$ 0.039 & 0.669 $\pm$ 0.036 \\
SP RoBERTa +  hard US
& 0.698 $\pm$ 0.012 & 0.732 $\pm$ 0.019 
& \bf{0.606} $\pm$ 0.244 & 0.686 $\pm$ 0.234
& \bf{0.661} $\pm$ 0.033 & \bf{0.742} $\pm$ 0.028 \\
\midrule
Zero-shot RoBERTa-NLI & 0.315 $\pm$ 0.000 & 0.382 $\pm$ 0.000
& 0.090 $\pm$ 0.000 & 0.110 $\pm$ 0.000 
& 0.065 $\pm$ 0.000 & 0.068 $\pm$ 0.000 \\
SP RoBERTa-NLI & 0.748 $\pm$ 0.026 & 0.801 $\pm$ 0.028 
& 0.669 $\pm$ 0.185 & \bf{0.758} $\pm$ 0.151 
& 0.700 $\pm$ 0.040 & 0.771 $\pm$ 0.031 \\
SP BERT-NLI  & 0.693 $\pm$ 0.017 & 0.738 $\pm$ 0.015 
& 0.624 $\pm$ 0.231 & 0.715 $\pm$ 0.197 
& 0.614 $\pm$ 0.035 & 0.695 $\pm$ 0.026 \\
SP BART-NLI & \bf{0.789} $\pm$ 0.024 & \bf{0.830} $\pm$ 0.030 
& \bf{0.673} $\pm$ 0.174 & 0.753 $\pm$ 0.143
& \bf{0.770} $\pm$ 0.039 & \bf{0.829} $\pm$ 0.034 \\
\bottomrule
\end{tabular}
}
\caption{Ablation study and task transferring: comparison on unseen intents. {\bf Top}: comparison of different contextualized encoders; {\bf middle}:  comparison of negative sampling strategies of intent sampling (IS) and utterance sampling (US); {\bf bottom}: task transferring from the MNLI dataset, using various fine-tuned models.}
\label{tab:table_ablation}
\end{threeparttable}
\end{table*}

\paragraph{Baselines} We use \textbf{SEG}\footnote{\url{https://github.com/fanolabs/0shot-classification}, unfortunately were unable to run the code and adopted the published results from the paper}, \textbf{RIDE}\footnote{\url{https://github.com/RIDE-SIGIR/GZS}}, \textbf{CTIR}\footnote{\url{https://github.com/PhoebusSi/CTIR}} as baselines, as they show the up-to-date top results on the three chosen datasets. For the RIDE model, we use the base model with a Positive-Unlabeled classifier, as it gives a significant improvement on the SGD and MultiWoZ datasets. We used Zero-Shot DNN and CapsNets along with CTIR, since these two encoders perform best on unseen intents \cite{si2020learning}. 

\paragraph{Evaluation metrics} commonly used for the task are accuracy (\textbf{Acc}) and \textbf{F1}. The F1 values are per class averages weighted with their respective support. Following previous works, we report results on {\bf seen} and {\bf unseen} intents separately. Evaluation for the test set {\bf overall} is presented in Appendix. We report averaged results along with standard deviation for ten runs of each experiment.

\paragraph{Results} of experiments are presented in Table~\ref{tab:all_results} (see Appendix for standard deviation estimation). Our approach SP RoBERTa, when used with intent labels and utterances only, shows significant improvement over the state-of-the-art on all three datasets, both on seen and unseen intents, by accuracy and F1 measures. The only exception is unseen intents of CLINC, where our approach under-performs in terms of accuracy of unseen intents recognition comparing to RIDE. At the same time, RIDE shows a lower recall score in this setup. So, our method is more stable and performs well even when the number of classes is high.

Similarly to other methods, our method recognizes seen intents better than unseen ones, reaching around 90\% of accuracy and F1 on the former. Next, with the help of lexicalized intent labels our approach yields even more significant improvement for all datasets. The gap between our approach and baselines becomes wider, reaching 14\% of accuracy on SGD's unseen intents and becoming closer to perfect detection on seen intents across all datasets. The difference between our base approach SP RoBERTa and its modification, relying on intent lexicalization, exceeds 7\% on unseen intents for SGD dataset and reaches 3\% on MultiWoZ ones. Notably, SP RoBERTa does not overfit on seen intents and achieves a consistent increase both on unseen and seen intents compared to previous works.


 \paragraph{Ablation study} We perform ablation studies for two parts of the SP RoBERTa approach and present the results for unseen intents in  Table~\ref{tab:table_ablation}. In all ablation experiments we use the SP approach with intent labels to diminish the effect of lexicalization.
 
First, we evaluate {\bf the choice of the contextualized encoder}, which is at the core of our approach (see the top part of Table~\ref{tab:table_ablation}). We choose between BERT$_{base}$, RoBERTa$_{base}$, its distilled version DistilRoBERTa, and TOD-BERT. BERT$_{base}$ provides poorer performance when compared to RoBERTa$_{base}$, which may be attributed to different pre-training setup. At the same time, TOD-BERT's scores are compatible with the ones of RoBERTa on two datasets, thus diminishing the importance of domain adaptation. A higher standard deviation, achieved for the MultiWoZ dataset, makes the results less reliable. The performance of DistilRoBERTa is almost on par with its teacher, RoBERTa, indicating that our approach can be used with a less computationally expensive model almost without sacrificing quality. 

Second, we experiment with the choice of {\bf negative sampling strategy} (see the middle part of Table~\ref{tab:table_ablation}), in which we can sample either random or hard negative examples for both intents and utterances. The overall trend shows that sampling hard examples improves over random sampling (by up to 6\% of accuracy for the SGD dataset).

\setlength{\textfloatsep}{0.1cm}
\begin{table*}[!ht] 
\centering
\begin{threeparttable}
\resizebox{\textwidth}{!}{%
\begin{tabular}{lcccccc}
\toprule
\multirow{3}{*}{\makecell{Intent \\description}} & 
\multicolumn{2}{c}{SGD} &
\multicolumn{2}{c}{MultiWoZ} &
\multicolumn{2}{c}{CLINC} \\
 \cmidrule{2-7}
 & Acc & F1 & Acc & F1 & Acc & F1 \\
\midrule
intent labels &0.687 $\pm$ 0.018 & 0.716 $\pm$ 0.016
& 0.594  $\pm$ 0.180 & 0.705 $\pm$0.157
& 0.639  $\pm$ 0.038 & 0.731 $\pm$ 0.028 \\
\midrule
d$_1$ templates & 0.750 $\pm$ 0.019 &0.805 $\pm$ 0.021
& 0.624 $\pm$ 0.231 & 0.722 $\pm$ 0.175
&0.692 $\pm$ 0.031&0.766 $\pm$ 0.028\\
d$_2$ templates & 0.752 $\pm$ 0.003 & 0.804 $\pm$ 0.006 
& 0.610 $\pm$ 0.219 & 0.713 $\pm$ 0.201
& 0.685 $\pm$ 0.035 & 0.756 $\pm$ 0.031\\
q$_1$ templates & 0.765 $\pm$ 0.019 &0.818 $\pm$ 0.021	
& 0.621 $\pm$ 0.208 & 0.727 $\pm$ 0.174
& 0.670 $\pm$ 0.034 & 0.747 $\pm$ 0.029 \\
q$_2$ templates &0.753 $\pm$ 0.026 & 0.807 $\pm$ 0.026
& 0.599 $\pm$ 0.212 & 0.702 $\pm$ 0.188
& 0.554 $\pm$ 0.054 & 0.620 $\pm$ 0.055\\

\bottomrule
\end{tabular}
}
\caption{Comparison of different lexicalization templates, improving the performance of SP RoBERTa. Metrics are reported on unseen intents only.  Each row corresponds to experiments with a single lexicalization template only, isolated from the others, i.e the row ``d$_1$ templates'' uses only the d$_1$ form. }
\label{tab:pattern_comparison}
\end{threeparttable}
\end{table*}

\setlength{\textfloatsep}{0.1cm}
\begin{table*}[!ht] 
\centering
\begin{threeparttable}
\resizebox{\textwidth}{!}{%
\begin{tabular}{ccccccccccccc}
\toprule
\multirow{3}{*}{Train data:  intent labels +} & 
\multicolumn{4}{c}{SGD} &
\multicolumn{4}{c}{MultiWoZ} &
\multicolumn{4}{c}{CLINC} \\
\cmidrule{2-13}
& \multicolumn{2}{c}{Unseen}&
\multicolumn{2}{c}{Seen}&
\multicolumn{2}{c}{Unseen}&
\multicolumn{2}{c}{Seen}&
\multicolumn{2}{c}{Unseen}&
\multicolumn{2}{c}{Seen} \\
\cmidrule{2-13}
 & Acc & F1 & Acc & F1 & Acc & F1 
 & Acc & F1 & Acc & F1 & Acc & F1 \\
 original utterances& 
0.687  &     0.716  &   0.916  &   0.922  &
0.594   & 0.705    &  0.903  & 0.912  & 
0.639  & 0.731  & 0.894  & 0.903 \\
synthetic utterances&
0.666  &        0.688 &   0.746  &      0.778  &
0.615  &        0.642  &   0.621  &      0.713  &
0.580  &        0.613  &   0.608  &      0.654  \\
\bottomrule
\end{tabular} }

\caption{Dataless classififcation. Metrics are reported on seen and unseen intents. Fine-tuning SP-Roberta on synthetic utterances (bottom) shows moderate decline, compared to training on real utterances (top).  }
\label{tab:dataless_results}
\end{threeparttable}
\end{table*}

\paragraph{Choice of lexicalization templates}  Table~\ref{tab:pattern_comparison} demonstrates the performance of SP RoBERTa with respect to the choice of lexicalization templates. Regardless of which template is used, the results achieved outperform SP RoBERTa with intent labels. The choice of lexicalization template slightly affects the performance. The gap between the best and the worst performing template across all datasets is about 2\%. The only exception is $q_2$, which drops the performance metrics for two datasets. In total, this indicates that our approach must use just any of the lexicalization templates, but which template exactly is chosen is not as important. What is more, there is no evidence that declarative templates should be preferred to questions or vice versa. 

Further adjustments of intent lexicalization templates and their derivation from the datasets seem a part of future research.  Other promising directions include using multiple lexicalized intent labels jointly to provide opportunities for off-the-shelf augmentation at the test and train times.

\paragraph{Task transferring}  results are presented in the  bottom part of Table~\ref{tab:table_ablation}. First, we experiment with zero-shot task transferring, using  RoBERTa-NLI to make predictions only, without any additional fine-tuning on intent recognition datasets. This experiment leads to almost random results, except for the SGD datasets, where the model reaches about 30\% correct prediction. 

However, models, pre-trained with MNLI and fine-tuned further for intent recognition, gain significant improvement up to 7\%. The improvement is even more notable in the performance of BART-NLI, which obtains the highest results, probably, because of the model's size.

\paragraph{Dataless classification} results are shown in Table~\ref{tab:dataless_results}. This experiment compares training on two datasets: (i) intent labels and original utterances, (ii) intent labels and synthetic utterances, achieved from paraphrasing lexicalized intent labels. In the latter case, the only available data is the set of seen intent labels, used as input to SP RoBERTa and for further paraphrasing. Surprisingly, the performance declines moderately: the metrics drop by up to 30\% for seen intents and up to 10\% for unseen intents. This indicates that a) the model learns more from the original data due to its higher diversity and variety; b) paraphrasing models can re-create some of the correlations from which the model learns.   

The series of experiments in transfer learning and dataless classification aims at real-life scenarios in which different parts of annotated data are available. First, in zero-shot transfer learning, we do not access training datasets at all (Table \ref{tab:all_results}, Zero-shot RoBERTA NLI). Second, in the dataless setup, we access only {\bf seen} intent labels, which we utilize both as class labels and as a source to create synthetic utterances (Table \ref{tab:dataless_results}). Thirdly, our main experiments consider both {\bf seen} intents and utterances available (Table \ref{tab:all_results}, SP RoBERTA).  In the second scenario, we were able to get good scores that are more or less close to the best-performing model. We believe efficient use of intent labels overall and to generate synthetic data, in particular, is an important direction for future research. 

\section{Analysis} \label{sec:discussion}

\paragraph{Error analysis} shows, that SP RoBERTa tends to confuse intents, which (i)  are assigned with semantically similar labels or  (ii) share a word. For example, an unseen intent \texttt{get\_train\_tickets} gets confused with the seen intent \texttt{find\_trains}. Similarly, pairs of seen intents \texttt{play\_media} and \texttt{play\_song} or \texttt{find\_home\_by\_area} and \texttt{search\_house} are hard to distinguish.

We checked whether errors in intent recognition are caused by utterances' surface or syntax features. Following observations hold for the SGD dataset. Utterances, which take the form of a question, are more likely to be classified correctly: 93\% of questions are assigned with correct intent labels, while there is a drop for declarative utterances, of which 90\% are recognized correctly. The model's performance is not affected by the frequency of the first words in the utterance. From 11360 utterances in the test set, 4962 starts with 3-grams, which occur more than 30 times. Of these utterances, 9\% are misclassified, while from the rest of utterances, which start with rarer words, 10\% are misclassified. The top-3 most frequent 3-grams at the beginning of an utterance are \textit{I want to}, \textit{I would like}, \textit{I need to}.

\paragraph{Stress test for NLI models} \cite{naik2018stress} is a  typology for the standard errors of sentence pair models, from which we picked several typical errors that can be easily checked without additional human annotation. We examine whether one of the following factors leads to an erroneous prediction: (i) word overlap between an intent label and an utterance; (ii) the length of an utterance; (iii) negation or double negation in an utterance; (iv) numbers, if used in an utterance. Additionally, we measured the semantic similarity between intent labels and user utterances through the SentenceBERT cosine function to check whether it impacts performance.

\begin{table}[!ht]
\centering
\begin{tabular}{llll}
\toprule
Test & Correct  & Incorrect \\
\midrule
\# overlapping tokens & 0.94 & 0.63 \\
\# tokens in utterance & 14.96 & 13.96 \\
\# digits in utterance & 0.31 & 0.23 \\
\# neg. words in utterance  & 0.03 & 0.02 \\ 
\midrule
Semantic similarity & 0.22 & 0.21 \\
\bottomrule
\end{tabular}
\caption{Stress test of SP RoBERTa predictions. An utterance is more likely to be correctly predicted if it shares at least one token with the intent labels.}
\label{tab:stress_test}
\end{table}

Table~\ref{tab:stress_test} displays the stress test results for one of the runs of SP RoBERTa, trained with q$_1$ template on the SGD dataset. This model shows reasonable performance, and its stress test results are similar to models trained with other templates. The results are averaged over the test set. An utterance gets more likely to be correctly predicted if it shares at least one token with the intent label. However, the semantic similarity between intent labels and utterances matters less and is relatively low for correct and incorrect predictions. Longer utterances or utterances, which contain digits, tend to get correctly classified more frequently. The latter may be attributed to the fact that numbers are important features to intents, related to doing something on particular dates and with a particular number of people, such as \texttt{search\_house},  \texttt{reserve\_restaurant} or \texttt{book\_appointment}.

\section{Conclusion}

Over the past years, there has been a trend of utilizing natural language descriptions for various tasks, ranging from dialog state tracking \cite{cao-zhang-2021-comparative}, named entity recognition \cite{li2020unified} to the most recent works in text classification employing Pattern-Exploiting Training (PET) \cite{schick2020exploiting}. The help of supervision, expressed in natural language, in most cases not only improves the performance but also enables exploration of real-life setups, such as few-shot or (generalized) zero-shot learning. Such methods' success is commonly attributed to the efficiency of pre-trained contextualized encoders, which comprise enough prior knowledge to relate the textual task descriptions with the text inputs to the model. 

Task-oriented dialogue assistants require the resource-safe ability to support emerging intents without re-training the intent recognition head from scratch. This problem lies well within the generalized zero-shot paradigm. To address it, we present a simple yet efficient approach based on sentence pair modeling, suited for the intent recognition datasets, in which each intent is equipped with a meaningful intent label. We show that we establish new state-of-the-art results using intent labels paired with user utterances as an input to a contextualized encoder and conducting simple binary classification. Besides, to turn intent labels into plausible sentences, better accepted by pre-trained models, we utilized an easy set of lexicalization templates. This heuristic yet alone gains further improvement, increasing the gap to previous best methods. Task transferring from other sentence pair modeling tasks leads to even better performance. 

However, our approach has a few limitations: it becomes resource-greedy as it requires to loop over all intents for a given utterance. Next, the intent labels may not be available or may take the form of numerical indices. The first limitation might be overcome by adopting efficient ranking algorithms from the Information Retrieval area. Abstractive summarization, applied to user utterances, might generate meaningful intent labels. These research questions open a few directions for future work.  


\section*{Acknowledgement}
Ekaterina Artemova is supported by the framework of the HSE University Basic Research Program.

\bibliography{custom}
\bibliographystyle{acl_natbib}

\newpage
\appendix




%




\section{Reproducibility Checklist}

\subsection{Code}
Our code is enclosed in this submission: \texttt{gzsl.zip}.

\subsection{Computing infrastructure}
Each experiment runs on a single NVIDIA V100 16Gb. The longest experiment was running for less than 2.5 hours.

\subsection{Datasets}
All used datasets are described in the paper. Preprocessing for SGD and MultiWoZ dataset includes (i) selecting utterances from dialogues where users express a new intent, (ii) cleaning uninformative short phrases like acknowledgments and greetings. Preprocessed datasets are also included in \texttt{gzsl.zip}
The SGD dataset is released under CC BY-SA 4.0 license.
The MultiWoZ dataset is released under Apache License 2.0
To the best of our knowledge the CLINC dataset is released under CC-BY-3.0 license.

\subsection{Randomness}
All experiments could be reproduced using the fixed set of seeds $\{11..20\}$.

\subsection{Evaluation metrics}
All used metrics and our motivation to use them are described in the main paper. Metrics and an evaluation script are implemented in our code.

\subsection{Models and hyperparameters}
Our sentence pair model consists of the contextualized encoder itself, a dropout, and a linear on top of the embedding for [CLS] token. All hyperparameters for the model are fixed in our submission configs. Transformer tokenizers use truncation for utterance and intent description to speed up execution time. Specified values for lexicalized and non-lexicalized setups are reported in \texttt{README.md}.

Batch size, learning rate, scheduler, warm-up steps ratio, and other experiment parameters are specified for each dataset and fixed in configs. We used the top 100 out-of-class similar utterances with a positive one as a threshold for hard negative sampling.

\subsection{Hyperparameter Search}
We performed hyperparameter search using the following grid for each dataset.
\begin{itemize}
    \item Learning rate: [$2e^{-5}$, $5e^{-5}$]
    \item Batch size: [8, 16]
    \item Warm up steps ratio: [0.10, 0.15]
    \item Utterance max length: [20, 30, 40]
    \item Negative samples k: [5, 7] 
\end{itemize}
For each hyperparameter configuration, we averaged the results over five runs.

\subsection{Acceptability evaluation} Lexicalized intent labels help to increase performance since they form more plausible sentences than raw intent labels. This observation can be confirmed by estimating the acceptability of a sentence. We evaluate the acceptability of intent labels and their lexicalized versions with several unsupervised measures, which aim to evaluate to which degree the sentence is likely to be produced \cite{lau2017grammaticality}.  We exploit the acceptability evaluation tool from \cite{lau2020furiously} with default settings.  Following acceptability measures have been used: $LP$ stands for unnormalized $\log$ probability of the sentence, estimated by a language model. $LP_{mean}$ and $LP_{pen}$ are differently normalized versions of $LP$ with respect to the sentence length. $LP_{norm}$ and $SLOR$ utilize additional normalization with unigram probabilities, computed over a large text corpus. In this experiment, BERT$_{large}$ is used as the default language model; unigram probabilities are pre-computed from bookcorpus-wikipedia. Higher acceptability scores stand for the higher likelihood of the sentence. Thus, more plausible and more natural-sounding sentences gain higher acceptability scores.   

We apply one of the lexicalization patterns to all intent labels, score each resulting sentence, and average the achieved scores.  Tables \ref{tab:acceptability_metrics_clinc}-\ref{tab:acceptability_metrics_multiwoz} present with the results of acceptability evaluation for the each dataset. As expected, the intent labels gain lower acceptability scores, while lexicalized patterns receive higher acceptability scores.   We may treat the acceptability of the pattern as a proxy to its performance since the $SLOR$ value of the poor performing  $q_2$ pattern is lower than for other patterns.

\onecolumn
\begin{table}[!ht]
\centering

\begin{tabular}{p{0.8cm}p{1cm}p{1cm}p{1cm}p{1cm}p{1cm}}

\toprule
{ID} &     $LP$ &  $LP_{mean}$ &  $LP_{pen}$ &  $LP_{norm}$ &  $SLOR$ \\
\midrule
labels & -34.58 &  -18.40 & -30.32 &   -1.84 & -8.44 \\
\midrule
d$_1$     & -43.16 &   -5.55 & -23.50 &   -0.74 &  1.97 \\
d$_2$      & -49.92 &   -5.68 & -25.60 &   -0.77 &  1.67 \\
\midrule
q$_1$     & -46.71 &   -5.31 & -23.93 &   -0.71 &  2.12 \\
q$_2$     & -43.86 &   -6.48 & -25.48 &   -0.87 &  0.98 \\
\bottomrule
\end{tabular}
\caption{Averaged acceptability scores, computed for the CLINC dataset. Rows stand for intent labels without any changes or lexicalized, using one of the patterns.  Higher acceptability scores mean that a sentence is more likely to be grammatical and sound natural. Intent labels less acceptable, while their lexicalized versions form plausible sentences.}
\label{tab:acceptability_metrics_clinc}

\end{table}

\begin{table}[!ht]
\centering
\begin{tabular}{p{0.8cm}p{1cm}p{1cm}p{1cm}p{1cm}p{1cm}}
\toprule
{ID} &     LP &  LP$_{mean}$ &  LP$_{pen}$ &  LP$_{norm}$ &  SLOR \\
\midrule
labels & -43.18 &  -18.45 & -36.31 &   -1.96 & -9.01 \\ \midrule
d$_1$     & -38.93 &   -5.45 & -22.08 &   -0.72 &  2.11 \\
d$_2$     & -43.00 &   -5.29 & -22.92 &   -0.71 &  2.09 \\ \midrule
q$_1$     & -41.91 &   -5.14 & -22.32 &   -0.69 &  2.31 \\
q$_2$     & -39.36 &   -6.44 & -23.93 &   -0.86 &  1.07 \\
\bottomrule
\end{tabular}

\caption{Acceptability measures, computed for the SGD dataset}
\label{tab:acceptability_metrics_sgd}
\end{table}

\begin{table}[!ht]
\centering
\begin{tabular}{p{0.8cm}p{1cm}p{1cm}p{1cm}p{1cm}p{1cm}}

\toprule
{ID} &     LP &  LP$_{mean}$ &  LP$_{pen}$ &  LP$_{norm}$ &  SLOR \\
\midrule
labels & -41.92 &  -20.96 & -37.06 &   -2.28 & -11.77 \\
\midrule

d$_1$    & -31.45 &   -4.49 & -18.07 &   -0.62 &   2.71 \\
d$_2$     & -33.90 &   -4.24 & -18.26 &   -0.60 &   2.83 \\
\midrule
q$_1$     & -33.73 &   -4.22 & -18.17 &   -0.59 &   2.93 \\
q$_2$     & -31.87 &   -5.31 & -19.62 &   -0.75 &   1.78 \\
\bottomrule
\end{tabular}

\caption{Acceptability measures, computed for the MultiWOZ dataset}
\label{tab:acceptability_metrics_multiwoz}

\end{table}

\begin{table*}[ht]

\resizebox{\textwidth}{!}{%
\begin{tabular}{ccccccc}
\toprule
\multirow{3}{*}{Method} & 
\multicolumn{2}{c}{Unseen} &
\multicolumn{2}{c}{Seen} &
\multicolumn{2}{c}{Overall} \\
 \cmidrule{2-7}
 & Acc & F1 & Acc & F1 & Acc & F1 \\
\midrule
SP RoBERTa + random IS &
0.687 $\pm$ 0.018 &     0.716 $\pm$ 0.016 &   0.916 $\pm$ 0.005 &   0.922 $\pm$ 0.004 &  0.884 $\pm$ 0.006 &  0.886 $\pm$ 0.005 \\
SP RoBERTa +  random US &
0.677 $\pm$ 0.017 &     0.707 $\pm$ 0.014 &   0.919 $\pm$ 0.005 &   0.932 $\pm$ 0.006 &  0.885 $\pm$ 0.005 &  0.893 $\pm$ 0.005 \\
SP RoBERTa +  hard IS &
0.741 $\pm$ 0.010 &     0.786 $\pm$ 0.017 &   0.884 $\pm$ 0.010 &   0.891 $\pm$ 0.012 &  0.864 $\pm$ 0.009 &  0.868 $\pm$ 0.010 \\
SP RoBERTa +  hard US
& 0.698 $\pm$ 0.012 & 0.732 $\pm$ 0.019 & 0.917  $\pm$ 0.003 
& 0.925 $\pm$ 0.003 &  0.887 $\pm$ 0.005 &  0.893 $\pm$ 0.008\\
\midrule
SP RoBERTa-NLI & 0.748 $\pm$ 0.026 &     0.801 $\pm$ 0.028 &   0.923 $\pm$ 0.004 &   0.929 $\pm$ 0.003 &  0.898 $\pm$ 0.005 &  0.905 $\pm$ 0.005 \\
SP BERT-NLI  & 0.693 $\pm$ 0.017 &     0.738 $\pm$ 0.015 &   0.918 $\pm$ 0.002 &   0.924 $\pm$ 0.001 &  0.886 $\pm$ 0.003 &  0.892 $\pm$ 0.002 \\
SP BART-NLI & 0.789 $\pm$ 0.024 &     0.830 $\pm$ 0.030 &   0.917 $\pm$ 0.000 &   0.924 $\pm$ 0.000 &  0.899 $\pm$ 0.003 &  0.907 $\pm$ 0.005 \\
\midrule
SP RoBERTa + d$_1$ patterns & 0.750 $\pm$ 0.019 &     0.805 $\pm$ 0.021 &   0.931 $\pm$ 0.006 &   0.934 $\pm$ 0.004 &  0.906 $\pm$ 0.004 &  0.909 $\pm$ 0.002\\
SP RoBERTa + d$_2$ patterns  & 0.752 $\pm$ 0.003 &     0.804 $\pm$ 0.006 &   0.927 $\pm$ 0.007 &   0.932 $\pm$ 0.004 &  0.902 $\pm$ 0.005 &  0.908 $\pm$ 0.003\\
SP RoBERTa + q$_1$ patterns & 0.765 $\pm$ 0.019 &     0.818 $\pm$ 0.021 &   0.922 $\pm$ 0.010 &   0.927 $\pm$ 0.010 &  0.900 $\pm$ 0.007 &  0.905 $\pm$ 0.007 \\
SP RoBERTa + q$_2$ patterns & 0.753 $\pm$ 0.026 &     0.807 $\pm$ 0.026 &   0.927 $\pm$ 0.005 &   0.931 $\pm$ 0.002 &  0.903 $\pm$ 0.004 &  0.908 $\pm$ 0.004\\
\bottomrule
\end{tabular}
}
\caption{Ablation study, task transferring and lexicalization patterns for SGD dataset. {\bf Top}: comparison of negative sampling strategies of intent sampling (IS) and utterance sampling (US); {\bf middle}: task transferring from the MNLI dataset, using various fine-tuned models; {\bf bottom}: Comparison of different lexicalization patterns, improving performance of SP RoBERTa.}
\label{tab:table_sgd}
\end{table*}

\begin{table*}[ht]
\centering
\resizebox{\textwidth}{!}{%
\begin{tabular}{ccccccc}
\toprule
\multirow{3}{*}{Method} & 
\multicolumn{2}{c}{Unseen} &
\multicolumn{2}{c}{Seen} &
\multicolumn{2}{c}{Overall} \\
 \cmidrule{2-7}
 & Acc & F1 & Acc & F1 & Acc & F1 \\
\midrule
SP RoBERTa + random IS & 0.594  $\pm$  0.180 & 0.705   $\pm$ 0.157 &  0.903 $\pm$ 0.055 & 0.912 $\pm$ 0.047 & 0.769 $\pm$ 0.082 & 0.767 $\pm$ 0.084\\
SP RoBERTa +  random US &
0.531 $\pm$ 0.218 &     0.632 $\pm$ 0.217 &   0.930 $\pm$ 0.036 &   0.938 $\pm$ 0.027 &  0.742 $\pm$ 0.096 &  0.730 $\pm$ 0.106 \\
SP RoBERTa +  hard IS &
0.561 $\pm$ 0.177 &     0.680 $\pm$ 0.136 &   0.937 $\pm$ 0.024 &   0.943 $\pm$ 0.016 &  0.771 $\pm$ 0.083 &  0.761 $\pm$ 0.091 \\
SP RoBERTa +  hard US &
0.606 $\pm$ 0.244 &        0.686 $\pm$ 0.234 &   0.903 $\pm$ 0.033 &      0.919 $\pm$ 0.030 &  0.764 $\pm$ 0.099 &     0.754 $\pm$ 0.108 \\
\midrule
SP RoBERTa-NLI & 0.669 $\pm$ 0.185 &     0.758 $\pm$ 0.151 &   0.943 $\pm$ 0.014 &   0.948 $\pm$ 0.012 &  0.808 $\pm$ 0.088 &  0.806 $\pm$ 0.089 \\
SP BERT-NLI  & 0.624 $\pm$ 0.231 &     0.715 $\pm$ 0.197 &   0.941 $\pm$ 0.011 &   0.948 $\pm$ 0.010 &  0.785 $\pm$ 0.103 &  0.782 $\pm$ 0.105 \\
SP BART-NLI & 0.673 $\pm$ 0.174 &     0.753 $\pm$ 0.143 &   0.946 $\pm$ 0.012 &   0.950 $\pm$ 0.010 &  0.820 $\pm$ 0.079 &  0.814 $\pm$ 0.086 \\
\midrule
SP RoBERTa + d$_1$ patterns & 0.624 $\pm$ 0.231 &     0.722 $\pm$ 0.175 &   0.941 $\pm$ 0.011 &   0.948 $\pm$ 0.010 &  0.785 $\pm$ 0.103 &  0.782 $\pm$ 0.105\\
SP RoBERTa + d$_2$ patterns  & 0.610 $\pm$ 0.219 &     0.713 $\pm$ 0.201 &   0.944 $\pm$ 0.013 &   0.948 $\pm$ 0.011 &  0.786 $\pm$ 0.095 &  0.781 $\pm$ 0.104\\
SP RoBERTa + q$_1$ patterns & 0.621 $\pm$ 0.208 &     0.727 $\pm$ 0.174 &   0.946 $\pm$ 0.010 &   0.949 $\pm$ 0.010 &  0.789 $\pm$ 0.097 &  0.786 $\pm$ 0.101 \\
SP RoBERTa + q$_2$ patterns & 0.599 $\pm$ 0.212 &     0.702 $\pm$ 0.188 &   0.943 $\pm$ 0.020 &   0.948 $\pm$ 0.015 &  0.778 $\pm$ 0.094 &  0.775 $\pm$ 0.097\\
\bottomrule
\end{tabular}
}
\caption{Ablation study, task transferring and lexicalization patterns for MultiWoZ dataset.}
\label{tab:table_multiwoz}
\end{table*}

\begin{table*}[ht]
\centering
\resizebox{\textwidth}{!}{%
\begin{tabular}{ccccccc}
\toprule
\multirow{3}{*}{Method} & 
\multicolumn{2}{c}{Unseen} &
\multicolumn{2}{c}{Seen} &
\multicolumn{2}{c}{Overall} \\
 \cmidrule{2-7}
 & Acc & F1 & Acc & F1 & Acc & F1 \\
\midrule
SP RoBERTa + random IS
& 0.639 $\pm$ 0.038 & 0.731 $\pm$ 0.028 & 0.894 $\pm$ 0.009 & 0.903 $\pm$ 0.010 & 0.768 $\pm$ 0.017 & 0.760 $\pm$ 0.017 \\
SP RoBERTa +  random US &
0.658 $\pm$ 0.043 &     0.735 $\pm$ 0.036 &   0.942 $\pm$ 0.007 &   0.903 $\pm$ 0.010 & 0.791 $\pm$ 0.024 &     0.816 $\pm$ 0.019 \\
SP RoBERTa +  hard IS &
0.590 $\pm$ 0.039 & 0.669 $\pm$ 0.036 & 0.881 $\pm$ 0.008 & 0.901 $\pm$ 0.010 & 0.763 $\pm$ 0.020 & 0.754 $\pm$ 0.018 \\
SP RoBERTa +  hard US &
0.661 $\pm$ 0.033 &        0.742 $\pm$ 0.028 &   0.946 $\pm$ 0.007 &      0.954 $\pm$ 0.005 &  0.794 $\pm$ 0.018 &     0.789 $\pm$ 0.020 \\
\midrule
SP RoBERTa-NLI & 0.700 $\pm$ 0.040 &     0.771 $\pm$ 0.031 &   0.950 $\pm$ 0.004 &   0.955 $\pm$ 0.003 &  0.817 $\pm$ 0.020 &     0.836 $\pm$ 0.015 \\
SP BERT-NLI  & 0.614 $\pm$ 0.035 &     0.695 $\pm$ 0.026 &   0.930 $\pm$ 0.007 &   0.938 $\pm$ 0.007 &  0.762 $\pm$ 0.020 &     0.791 $\pm$ 0.018 \\
SP BART-NLI & 0.770 $\pm$ 0.039 & 0.829 $\pm$ 0.034
& 0.973 $\pm$ 0.003 & 0.976 $\pm$ 0.002
& 0.865 $\pm$ 0.022 & 0.862 $\pm$ 0.024 \\
\midrule
SP RoBERTa + d$_1$ patterns & 0.692 $\pm$ 0.031 &     0.766 $\pm$ 0.028 &   0.927 $\pm$ 0.009 &   0.931 $\pm$ 0.008 &  0.802 $\pm$ 0.018 &     0.817 $\pm$ 0.015\\
SP RoBERTa + d$_2$ patterns  & 0.685 $\pm$ 0.035 &     0.756 $\pm$ 0.031 &   0.923 $\pm$ 0.014 &   0.928 $\pm$ 0.012 &  0.796 $\pm$ 0.024 &     0.812 $\pm$ 0.021\\
SP RoBERTa + q$_1$ patterns & 0.670 $\pm$ 0.034 &     0.747 $\pm$ 0.029 &   0.925 $\pm$ 0.010 &   0.930 $\pm$ 0.009 &  0.789 $\pm$ 0.019 &     0.808 $\pm$ 0.015 \\
SP RoBERTa + q$_2$ patterns & 0.554 $\pm$ 0.054 &     0.620 $\pm$ 0.055 &   0.919 $\pm$ 0.008 &   0.921 $\pm$ 0.009 &  0.725 $\pm$ 0.029 &     0.752 $\pm$ 0.022\\
\bottomrule
\end{tabular}
}
\caption{Ablation study, task transferring and lexicalization patterns for CLINC dataset.}
\label{tab:table_clinc}
\end{table*}

\begin{sidewaystable*}[!ht] 
\centering
\begin{threeparttable}
\resizebox{\textwidth}{!}{%
\begin{tabular}{ccccccccccccc}
\toprule
\multirow{3}{*}{Train data} & 
\multicolumn{4}{c}{SGD} &
\multicolumn{4}{c}{MultiWoZ} &
\multicolumn{4}{c}{CLINC} \\
\cmidrule{2-13}
& \multicolumn{2}{c}{Unseen}&
\multicolumn{2}{c}{Seen}&
\multicolumn{2}{c}{Unseen}&
\multicolumn{2}{c}{Seen}&
\multicolumn{2}{c}{Unseen}&
\multicolumn{2}{c}{Seen} \\
\cmidrule{2-13}
 & Acc & F1 & Acc & F1 & Acc & F1 
 & Acc & F1 & Acc & F1 & Acc & F1 \\
\makecell{intent labels +\\ original utterances}& 
0.687 $\pm$ 0.018 &     0.716 $\pm$ 0.016 &   0.916 $\pm$ 0.005 &   0.922 $\pm$ 0.004 &
0.594  $\pm$  0.180 & 0.705   $\pm$ 0.157 &  0.903 $\pm$ 0.055 & 0.912 $\pm$ 0.047 & 
0.639 $\pm$ 0.038 & 0.731 $\pm$ 0.028 & 0.894 $\pm$ 0.009 & 0.903 $\pm$ 0.010 \\
\makecell{intent labels +\\ synthetic utterances}&
0.666 $\pm$ 0.019 &        0.688 $\pm$ 0.020 &   0.746 $\pm$ 0.014 &      0.778 $\pm$ 0.014 &
0.615 $\pm$ 0.138 &        0.642 $\pm$ 0.090 &   0.621 $\pm$ 0.101 &      0.713 $\pm$ 0.084 &
0.580 $\pm$ 0.045 &        0.613 $\pm$ 0.040 &   0.608 $\pm$ 0.016 &      0.654 $\pm$ 0.009 \\
\bottomrule
\end{tabular}
}
\caption{Dataless classififcation. Metrics are reported on seen and unseen intents. Fine-tuning SP-Roberta on synthetic utterances (bottom) shows moderate decline, compared to training on real utterances (top).  }
\label{tab:dataless_results_ext}
\end{threeparttable}

\bigskip\bigskip\bigskip\bigskip

\begin{threeparttable}
\resizebox{\textwidth}{!}{%
\begin{tabular}{ccccccccccccc}
\toprule
\multirow{3}{*}{Method} & 
\multicolumn{4}{c}{SGD} &
\multicolumn{4}{c}{MultiWoZ} &
\multicolumn{4}{c}{CLINC} \\
\cmidrule{2-13}
& \multicolumn{2}{c}{Unseen}&
\multicolumn{2}{c}{Seen}&
\multicolumn{2}{c}{Unseen}&
\multicolumn{2}{c}{Seen}&
\multicolumn{2}{c}{Unseen}&
\multicolumn{2}{c}{Seen} \\
\cmidrule{2-13}
 & Acc & F1 & Acc & F1 & Acc & F1 
 & Acc & F1 & Acc & F1 & Acc & F1 \\
\midrule
SEG& 0.372 & 0.403 & 0.613 & 0.636 & 0.371 & 0.414 & 0.652 & 0.646&
-& -& -& - \\
\midrule
RIDE+PU& 0.590 & 0.573 & 0.832 & 0.830 & 0.569 & 0.521 & 0.884 & 0.885 &
0.798 & 0.573 & 0.908 & 0.912 \\
\midrule
ZSDNN + CTIR&
    0.603$\pm$ 0.002 &        0.580$\pm$ 0.003 &   0.809$\pm$ 0.006 &      0.878$\pm$ 0.014 &
    0.468$\pm$ 0.185 &        0.437$\pm$ 0.176 &   0.827$\pm$ 0.022 &      0.892$\pm$ 0.035  &
    0.561$\pm$ 0.059 &        0.493$\pm$ 0.054 &   0.904$\pm$ 0.031 &      0.871$\pm$ 0.026 \\
\midrule
CapsNet + CTIR & 
    0.567 $\pm$ 0.017 &        0.507 $\pm$ 0.026 &   0.897 $\pm$ 0.010 &      0.912 $\pm$ 0.009 &
    0.481 $\pm$ 0.174 &        0.404 $\pm$ 0.243 &   0.903 $\pm$ 0.017 &      0.906 $\pm$ 0.026 &
    0.530 $\pm$ 0.049 &        0.572 $\pm$ 0.033 &   0.866 $\pm$ 0.014 &      0.883 $\pm$ 0.020 \\
\midrule
SP RoBERTa (ours)& 
0.698 $\pm$ 0.012 & 0.732 $\pm$ 0.019 & 0.917  $\pm$ 0.003 & 0.925 $\pm$ 0.003
& 0.606 $\pm$ 0.244 &        0.686 $\pm$ 0.234 &   0.903 $\pm$ 0.033 &      0.919 $\pm$ 0.030 
& 0.661 $\pm$ 0.033 &        0.742 $\pm$ 0.028 &   \bf{0.946 $\pm$ 0.007} &      \bf{0.954 $\pm$ 0.005}\\
\midrule
\makecell{SP RoBERTa\\ + patterns (ours)}& 
\bf{0.750 $\pm$ 0.019} &     \bf{0.805 $\pm$ 0.021} &   \bf{0.931 $\pm$ 0.006} &   \bf{0.934 $\pm$ 0.004} &
\bf{0.624 $\pm$ 0.231}&	\bf{0.722 $\pm$ 0.175}&	\bf{0.941 $\pm$ 0.011}&	\bf{0.948 $\pm$ 0.010}&
0.692 $\pm$ 0.031 &     \bf{0.766 $\pm$ 0.028} &   0.927 $\pm$ 0.009 &   0.931 $\pm$ 0.008\\
\bottomrule
\end{tabular}
}
\caption{Comparison of different methods. SP stands for Sentence Pair modeling approach. SP RoBERTa (ours) shows consistent improvements of F1 across all datasets for seen and unseen intents. The usage of lexicalized patterns improves performance. }

\label{tab:all_results_ext}
\end{threeparttable}
\end{sidewaystable*}


\end{document}